%% file: main.tex
\documentclass{article} 
\usepackage{iclr2021_conference,times}

\input{math_commands.tex}

\usepackage{hyperref}
\usepackage{url}
\usepackage{graphicx}
\usepackage{svg}
\usepackage{tabularx}
\usepackage{booktabs}
\usepackage{placeins}
\usepackage{enumitem}
\usepackage{xcolor}    
\title{Wan-Animate: Unified Character Animation and Replacement with Holistic Replication}

\author{HumanAIGC Team\\
Tongyi Lab, Alibaba} 

\iclrfinalcopy 
\makeatletter
\renewcommand{\headrulewidth}{0pt} 
\begin{document}

\maketitle

\begin{abstract}
We introduce \textit{Wan-Animate}, a unified framework for character animation and replacement. Given a character image and a reference video, \textit{Wan-Animate} can animate the character by precisely replicating the expressions and movements of the character in the video to generate high-fidelity character videos. Alternatively, it can integrate the animated character into the reference video to replace the original character, replicating the scene's lighting and color tone to achieve seamless environmental integration. \textit{Wan-Animate} is built upon the Wan model. To adapt it for character animation tasks, we employ a modified input paradigm to differentiate between reference conditions and regions for generation. This design unifies multiple tasks into a common symbolic representation. We use spatially-aligned skeleton signals to replicate body motion and implicit facial features extracted from source images to reenact expressions, enabling the generation of character videos with high controllability and expressiveness. Furthermore, to enhance environmental integration during character replacement, we develop an auxiliary Relighting LoRA. This module preserves the character's appearance consistency while applying the appropriate environmental lighting and color tone. Experimental results demonstrate that \textit{Wan-Animate} achieves state-of-the-art performance. We are committed to open-sourcing the model weights and its source code. Project Page: {\color{blue}\url{https://humanaigc.github.io/wan-animate/}}
\end{abstract}

\input{content/introduction}

\input{content/related_works}

\input{content/model_arch}

\input{content/implementation}

\input{content/experiments}

\section{Conclusion}
This paper introduces \textit{Wan-Animate}, a state-of-the-art method for character animation and replacement. \textit{Wan-Animate} supports two core functionalities:
Character Animation: Given a reference video and a character image, it drives the character image with the motion from the video to generate a new animation.
Character Replacement: Given a reference video and a character image, it replaces the character in the video with the new one.
We design a modified input paradigm that unifies these diverse input forms, making the training process more efficient. \textit{Wan-Animate} achieves precise reenactment of both facial expressions and body motions. For signal injection, we disentangle motion and expression. Motion signals are integrated with the input noise latents via spatially-aligned fusion, while expression signals are injected via attention using implicit features extracted from the facial image. Furthermore, for character replacement, we have designed an auxiliary LoRA module that enables the model to better achieve lighting and color tone consistency between the character and the new environment.
The performance of \textit{Wan-Animate} surpasses that of current open-source and closed-source algorithms. We will open-source \textit{Wan-Animate} to contribute to the further iteration and application of this technology.

\section{Contributors}
All contributors are listed in alphabetical order by their last names.

\begin{itemize}[nosep,label={},leftmargin=*]
    \bfseries
    \color{blue}
    \item Gang Cheng, Xin Gao, Li Hu, Siqi Hu, Mingyang Huang, Chaonan Ji, Ju Li, Dechao Meng, Jinwei Qi, Penchong Qiao, Zhen Shen, Yafei Song, Ke Sun, Linrui Tian, Feng Wang, Guangyuan Wang, Qi Wang, Zhongjian Wang, Jiayu Xiao, Sheng Xu, Bang Zhang, Peng Zhang, Xindi Zhang, Zhe Zhang, Jingren Zhou, Lian Zhuo
    
\end{itemize}


\FloatBarrier

\bibliography{iclr2021_conference}
\bibliographystyle{iclr2021_conference}


\end{document}

%% file: math_commands.tex
\usepackage{amsmath,amsfonts,bm}

\def\eqref#1{equation~\ref{#1}}

\def\1{\bm{1}}

\DeclareMathAlphabet{\mathsfit}{\encodingdefault}{\sfdefault}{m}{sl}
\SetMathAlphabet{\mathsfit}{bold}{\encodingdefault}{\sfdefault}{bx}{n}



%% file: content/introduction.tex
\section{Introduction}
\label{intro}

Character image animation is a significant research area that has achieved remarkable progress, driven by advancements in visual generative techniques, particularly the application of diffusion models. This technology holds substantial potential value, with wide-ranging applications in filmmaking, advertising, and the creation of digital avatars.
Previous works, such as \cite{aa,champ,stableanimator}, have thoroughly explored diffusion-based architectures for character image animation, introducing significant enhancements in consistency and controllability. Furthermore, some studies \cite{aa2,mimo,anchorcrafter,dreamactorh1} have incorporated environmental information, extending the generative capabilities to more versatile tasks like character replacement and human-object interaction synthesis.
More recently, following the introduction of Sora \cite{sora} by OpenAI, video generation leveraging the Diffusion Transformer \cite{dit} architecture has undergone explosive development. The emergence of numerous open-source works \cite{hunyuanvideo, wan2025} has catalyzed parallel advancements in related sub-tasks. Consequently, DiT-based character image animation \cite{humandit, realisdancedit, dreamactorm1} has also garnered considerable research interest. By capitalizing on the capabilities of pre-trained video foundation models, the realism and temporal coherence of the generated character videos have been substantially improved.

However, a critical gap remains: no existing framework provides a holistic solution for high-fidelity character animation that unifies the control of motion, expression, and environment interaction. Within the open-source domain, existing works on character image animation exhibit significant shortcomings in both performance and completeness. The majority of open-source contributions, such as \cite{champ,mimicmotion,stableanimator,unianimate}, are designed based on UNet-based foundation models (e.g., SD \cite{ldm}, SVD \cite{svd}), and their results lag considerably behind the current state-of-the-art. 
While some DiT-based open-source projects \cite{realisdancedit,unianimatedit} focus on motion control, they fall short in holistically replicating expressive facial dynamics in conjunction with body movements. Furthermore, there is a scarcity of dedicated open-source methods for integrating character animation with environmental contexts (i.e., character replacement). Although some video generation methods \cite{vace} can approximate this functionality, they typically suffer from issues with consistency and usability. These unresolved challenges collectively hinder the continuous development and innovation of character image animation within the open-source community.

\begin{figure*}[tb]
  \centering
  \includegraphics[width=1.0\textwidth]{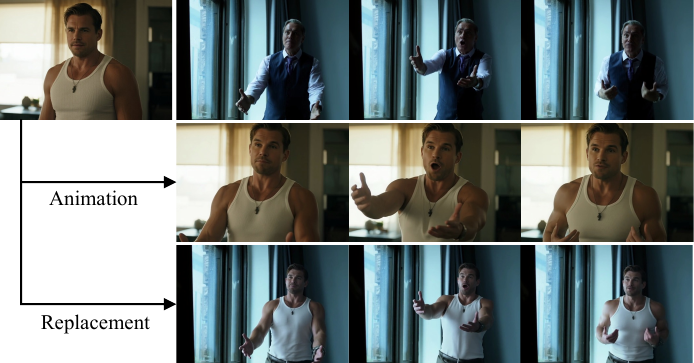}
  \caption{Given a character image and a reference video, \textit{Wan-Animate} supports two core functionalities. In the first, which we term "Animation," it reenacts the motion and expression of the character in the reference video to animate the static source image. In the second, termed "Replacement," it substitutes the character in the reference video with the source identity, ensuring seamless integration with the environment.
  }
  \label{fig:f1}
\end{figure*}

To address the aforementioned issues, this report introduces a unified framework for Character Animation and Replacement, termed \textit{Wan-Animate}, which achieves holistic replication with high-fidelity results. As illustrate in Figure \ref{fig:f1}, given a character image and a reference video, \textit{Wan-Animate} can accurately replicate the facial expressions and body movements from the reference to animate the character, generating a realistic character video. Concurrently, \textit{Wan-Animate} supports character replacement, enabling the animated character to be integrated into the reference video to replace the original character. This process also replicates the video's lighting and color tone, achieving a seamless fusion of the character and the environment.

Methodologically, \textit{Wan-Animate} is built upon the Wan-I2V model as its foundation and is enhanced through post-training with additional control conditions. Compared to Wan-I2V, \textit{Wan-Animate} features a modified input definition tailored for the demands of character animation. We employ a modified input paradigm to differentiate between reference conditions and regions designated for generation, which in turn guides the injection of corresponding latents. This design unifies reference image injection, temporal frame guidance, and the mode selection between full-frame generation and character replacement into a common symbolic representation. Crucially, this approach preserves the original input structure of Wan-I2V, thereby minimizing distributional shift during post-training.
To achieve holistic character control, we decouple the control signals into body motion and facial expressions. For body motion, we adopt a skeleton-based representation to balance accuracy and generality. As this signal is spatially aligned, it is injected by being added to the initial noise latents. For expression replication, we directly use the original face images from the reference video as the driving signal to preserve maximum detail. These face images are encoded into latent vectors to disentangle expression information from identity attributes. These latents are then temporally compressed to align with the video latents and are injected into the model via cross-attention. This joint control strategy demonstrates high robustness and precision.
When performing character replacement, we develop an auxiliary Relighting LoRA to enhance the consistency between character and environment. After the base model ensures the consistent transfer of the character's appearance, this module applies appropriate environmental lighting and color tones, resulting in a more seamless integration of the replaced character into the video.

We are committed to releasing the entire \textit{Wan-Animate} framework to the public, encompassing the model weights and the complete pipeline. Experimental results demonstrate that \textit{Wan-Animate} possesses excellent and versatile capabilities. It not only animates characters to generate expressive videos but also generalizes well to a wide range of humanoid characters, demonstrating strong robustness across various scenarios such as portraits, half-body, and full-body shots. Furthermore, it exhibits a competitive advantage in quality even when compared to several closed-source commercial products. We hope that the release of \textit{Wan-Animate} will make a significant contribution to accelerating the development of character image animation. We also aim to empower developers of all levels with access to a high-caliber model, enabling them to build diverse applications, inspire novel product paradigms, and ultimately facilitate the technology's real-world deployment.

%% file: content/related_works.tex
\section{Related Works}

\textbf{Character Image Animation. }
Earlier works in image animation primarily focused on warping-based feature representations and GAN-based architectures \cite{fomm,tpsmm,mraa}. In contrast, recent approaches \cite{aa,magicanimate,magicpose,champ,stableanimator} have shifted to designing architectures based on diffusion models \cite{ddpm}, which has led to significant improvements in performance. For instance, Animate Anyone \cite{aa} utilizes a ReferenceNet structure to inject the character's appearance features, achieving excellent results in consistency preservation. For temporal modeling, it employs temporal layers inspired by AnimateDiff \cite{animatediff}, which are embedded within the Stable Diffusion \cite{ldm} architecture. This design paradigm has been adopted by many subsequent methods \cite{champ, stableanimator}.
However, as a foundation model primarily for image generation, Stable Diffusion lacks inherent temporal knowledge. To address this limitation, some works \cite{unianimate, mimicmotion} have started building upon video foundation models. Since these models already possess knowledge of inter-frame consistency and continuity from their pre-training, the resulting image animation architectures can be more streamlined. More recently, with the dramatic improvement in video generation capabilities brought by DiT-based models \cite{hunyuanvideo,wan2025,yang2024cogvideox}, their application to image animation \cite{realisdancedit,dreamactorm1} has led to substantial enhancements in the realism and temporal continuity of generated characters.
Correspondingly, our \textit{Wan-Animate} is also built upon the open-source model Wan2.1, fully leveraging its robust pre-trained knowledge to ensure high-quality visual generation from the outset.

Beyond the generation of the character itself, some works also investigate the fusion of the animated character with its surrounding environment or with objects it interacts with. For instance, \cite{mimo} takes an additional background image as input, enabling the generated character video to feature a specified scene. Another approach, proposed by \cite{aa2}, utilizes a masking mechanism to differentiate between the background environment and interactive objects, which results in generated characters that are highly compatible with the environment. Conceptually, these methods can be adapted for the task of video character replacement.
Furthermore, a related line of research focuses on generating videos of human-object interactions \cite{anchorcrafter,dreamactorh1}. These methods aim to generate not only a moving character but also a co-moving object that is animated in a physically consistent and synchronized manner.

\textbf{Facial Animation. }
The field of facial animation \cite{lia,liveportrait} has also benefited significantly from the application of diffusion models, achieving remarkable advancements. Drawing inspiration from pose-guided human animation, some early works \cite{aniportrait,emoji,mimicmotion,stableanimator} employed facial landmarks as control signals to generate expressions, with their architectural designs closely following the human animation paradigm. However, compared to body poses, facial landmarks can lose fine-grained details during extraction, which compromises the expressiveness of the resulting animations and makes it difficult to synthesize subtle expressions. Moreover, in cross-identity application scenarios, dense facial landmarks demand high precision in signal retargeting, posing a significant challenge when driving diverse character identities.
More recently, some methods \cite{vasa,emoportraits,float,xportrait,xportrait2,hunyuanportrait,dreamactorm1} have begun to move away from manually defined motion signals, instead using the raw source images to extract implicit representations as control inputs. This approach has led to substantial improvements in both expressiveness and generality. 

\textbf{Video Generation. }
Since the release of Sora \cite{sora} by OpenAI, DiT-based \cite{dit} approaches have gradually supplanted UNet-based ones, becoming the mainstream technical route in video generation research. The adoption of a pure Transformer architecture allows model parameters to be scaled up significantly. Coupled with the expansion of training data, this has resulted in a qualitative leap in video generation quality. To fuse multi-modal features within a Transformer, videos are tokenized into discrete sequences. Specifically, a 3D VAE \cite{vae} is first used to compress the video in both spatial and temporal dimensions, drastically reducing feature length. The compressed representation is then discretized through a patchifying process. Most recent DiT-based video generation models \cite{hunyuanvideo,wan2025,yang2024cogvideox} follow this pipeline.
As the technology has matured, a growing number of foundation models for video generation have been open-sourced, with HunyuanVideo and Wan being particularly representative works. These have spurred a surge of follow-up research and applications in video generation. Consequently, the field of character image animation has also benefited from the advancements in video foundation models, with its performance gradually aligning with the capabilities of general video generation.

%% file: content/model_arch.tex
\section{Model Design and Architecture}

\begin{figure*}[tb]
  \centering
  \includegraphics[width=1.0\textwidth]{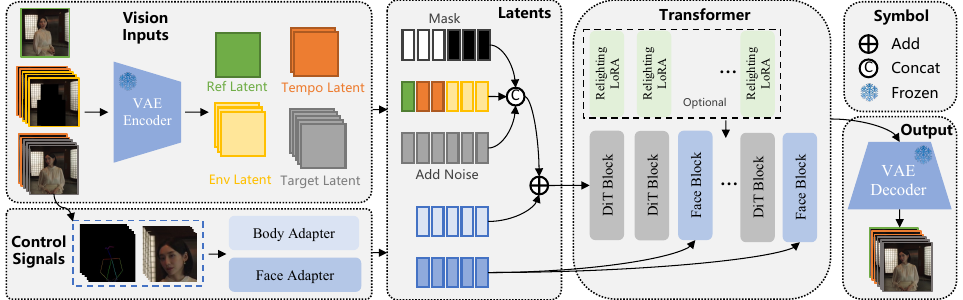}
  \caption{Overview of \textit{Wan-Animate}, which is built upon Wan-I2V. We modify its input formulation to unify reference image input, temporal frame guidance, and environmental information (for dual-mode compatibility) under a common symbolic representation. For body motion control, we use skeleton signals that are merged via spatial alignment. For facial expression control, we leverage implicit features extracted from face images as the driving signal. Additionally, for character replacement, we train an auxiliary Relighting LoRA to enhance the character's integration with the new environment.
  }
  \label{fig:arch}
\end{figure*}

\subsection{Task Definition}
\textit{Wan-Animate} features two core functionalities: Animation Mode and Replacement Mode.
In Animation Mode, the character from a source image is animated according to the motion of the character in a reference video, while the background from the source image is preserved. This process is analogous to an Image-to-Video (I2V) synthesis task.
In Replacement Mode, the character from the source image is driven by the same reference motion but is then integrated into the environment of the reference video. This effectively replaces the original subject, a task that corresponds to Video-to-Video (V2V) translation.
The common objective of both modes is to accurately replicate the motion and facial expressions from the reference character. The key distinction lies in the source of the final video's background: in Animation Mode, it is derived from the source image, whereas in Replacement Mode, it is inherited from the reference video. \textit{Wan-Animate} unifies both modes within a single, jointly trained model, with the exception of the Relighting LoRA which is specific to the Replacement Mode. By making minor adjustments to the input format, the model can generate outputs in the desired mode. The overall architecture is shown in Figure \ref{fig:arch}.

\subsection{Inputs Formulation}

\textit{Wan-Animate} leverages Wan-I2V as its foundational architecture. The input to Wan-I2V consists of three components: noise latent, conditional latent, and binary mask. Since the I2V task is defined as generating a video from a given image as the first frame, the conditional latent is constructed by concatenating the given image with zero-filled frames along the temporal dimension. The binary mask, which shares the same spatial and temporal dimensions as the conditional latent, uses a value of 1 to denote preserved frames and 0 for frames to be generated. For I2V, only the mask for the first frame is set to 1.
However, character image animation imposes different requirements on the input paradigm. Firstly, unlike the I2V setup where the image serves as the starting frame, our task requires a character image to act as a consistent appearance reference. The content of the generated video is dictated by driving signals, not initiated from the character image itself. Secondly, to enable animation of arbitrary length, the generation of subsequent segments must be conditioned on the final frame(s) of the preceding segment. This provides temporal guidance and ensures continuity, facilitating the synthesis of long videos. Third, we aim to unify the Animation Mode and Replacement Mode into a single model through a compatible representation, thereby reducing redundant training efforts.
Therefore, to accommodate these unique demands, \textit{Wan-Animate} introduces a modified input paradigm based on the original Wan-I2V formulation.

\textbf{Reference Formulation. }
Given a reference character image, we first encode it into a dedicated reference latent using the Wan-VAE. To fully leverage the inter-frame consistency capabilities pre-trained in the Wan model, the reference latent is concatenated with the conditional latents along the temporal dimension (with the binary mask set to 1). This concatenation serves as the primary mechanism for injecting the character's appearance. 
To accommodate the temporal guidance required for long video synthesis, we randomly select the first few latents from the target sequence to serve as temporal latents. For these selected latents, their corresponding ground-truth values are used as the condition latents, and the associated binary mask is set to 1 across the entire frame. This enables the model to generate temporally coherent videos guided by these temporal frames.
We employ a probabilistic training strategy where temporal latents are used only with a certain probability. This approach ensures the model learns to balance its generative capabilities across different conditional inputs.
Notably, the denoising process generates a complete output sequence, including the portions for the references. The resulting frames that correspond to these references are subsequently discarded.

\textbf{Environment Formulation. }
In Animation Mode, the conditional frames corresponding to the target frames are zero-filled, and their associated binary mask is set entirely to 0. Consequently, \textit{Wan-Animate} generates the character video while preserving the background from the given reference image, a process analogous to the standard I2V mode.
In Replacement Mode, we first segment the character from the reference video. Following the mask formulation strategy from \cite{aa2}, we then generate environment images by zeroing out the segmented subject region. This environment image serves as the content for the condition frames. Correspondingly, the binary mask is set to 1 for the environment regions and 0 for the subject region. As a result, \textit{Wan-Animate} only generates content within the mask-zeroed areas, thereby preserving the original background of the reference video.

In summary, the input paradigm of \textit{Wan-Animate}, while adapted for new tasks, fundamentally inherits the core philosophy of Wan-I2V. This design elegantly accommodates the diverse conditional requirements of character animation and supports dual generative modes. This adaptability allows the model to be fine-tuned rapidly and effectively during post-training, leading to strong empirical results.

\subsection{Control Signals}

\textbf{Body Control. }
Prior research has demonstrated the effectiveness of spatially-aligned signals for guiding human video generation. In terms of current technical approaches, there are two primary types of body control signals: 2D skeleton-based representations and rendered images from 3D SMPL \cite{smpl}. The skeleton-based approach offers better generality, particularly for non-humanoid characters with unconventional shapes, demonstrating greater robustness. However, it faces challenges in representing complex motions due to spatial ambiguity and is susceptible to issues like missing or erroneous keypoints. Conversely, SMPL, as a 3D signal, provides a more accurate representation of inter-limb relationships in complex poses but may lack precision for extremity positions and has poor capture capability for non-human characters. Additionally, rendered SMPL images contain the character's shape information. This can cause the model to rely on the shape cues embedded within the motion signal, which complicates the learning of identity consistency, especially if the SMPL shape is not accurate.
After careful consideration, we adopt the skeleton-based representation for body control, as it better caters to the majority of mainstream use cases. In our implementation, the skeleton for the character in the target frames is extracted using VitPose \cite{vitpose} to generate pose frames. In our design of Body Adapter, these pose frames are compressed by Wan-VAE to align spatially and temporally with the target latents. We use a projection layer to patchify the pose latents and add them to the patchified noise latents. Crucially, the reference latent is not injected with pose information. This design choice serves to temporally differentiate the reference latent from the target latents.

\begin{figure*}[tb]
  \centering
  \includegraphics[width=0.8\textwidth]{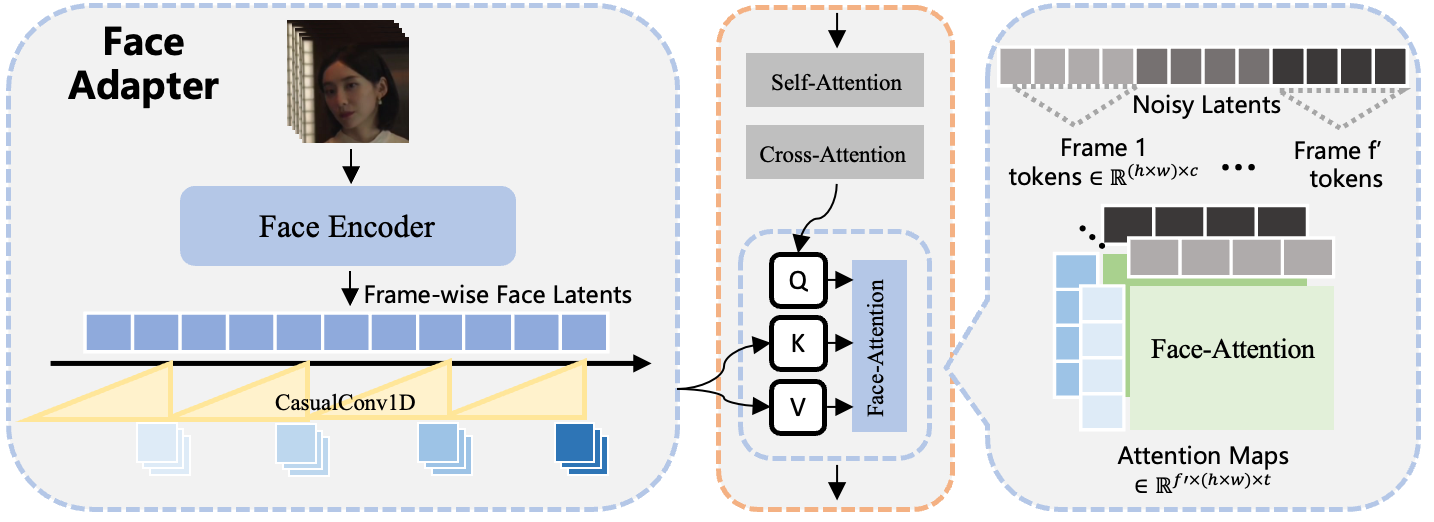}
  \caption{Face images are encoded into frame-wise implicit latents, which are then temporally aligned with the DiT latents. These features are injected via a cross-attention mechanism that operates within each corresponding temporal segment.
  }
  \label{fig:face}
\end{figure*}

\textbf{Face Control. }
A straightforward approach would be to use facial landmarks as a spatially-aligned signal for driving facial animation, similar to body control. However, this method suffers from a loss of fine-grained detail during landmark extraction, making it difficult to fully replicate the expressiveness of the character from the reference video. Moreover, as dense signals, facial landmarks demand high precision; otherwise, they can severely compromise identity consistency, especially in cross-identity scenarios involving significant facial shape disparities.
In contrast, we avoid manually defined facial signals and instead use the raw facial image directly as the driving input. During training, we leverage the character's skeletal information to locate and crop the facial region from the driving image. Since our training is self-supervised, it is crucial to disentangle identity information from expression information when extracting facial features. This prevents the model from using identity cues to guide generation, which could lead to identity leakage.
We employ two primary strategies to address this challenge:
1) We spatially compress the facial image into a 1D latent, which reduces the storage of low-level, identity-specific information.
2) During training, we apply a suite of data augmentations to the facial image, including scaling, color jittering, and random noise. This introduces deliberate discrepancies between the augmented face and the target face, discouraging the model from overfitting to identity features.

Architecturally, in Face Adapter, we adopt an encoder structure identical to that of \cite{lia} to extract features from each face image. We also employ Linear Motion Decomposition to orthogonalize these features, which facilitates a better disentanglement of expression information. Input face images are resized to $512\times512$, and each frame is compressed into a latent vector. As shown in Figure \ref{fig:face}, we use a stack of 1D causal convolutional layers to temporally downsample the face latents, aligning their sequence length with that of the noise latents.
The aligned face latents are then injected into dedicated "Face Blocks" within the Transformer. Feature fusion is achieved via a temporally-aligned cross-attention mechanism, where the attention computation is confined to the corresponding set of tokens at each timestep. To reduce the computational load, we opt to inject face information only into specific layers of the DiT network. Empirically, we perform this injection every 5 layers in the 40-layer Wan-14B model, resulting in a total of 8 injection layers.

\begin{figure*}[tb]
  \centering
  \includegraphics[width=1.0\textwidth]{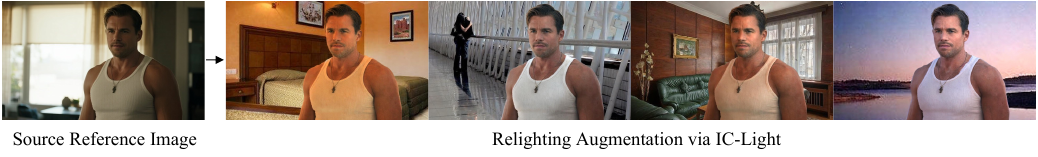}
  \caption{Examples of data augmentation using IC-Light.
  }
  \label{fig:iclight}
\end{figure*}

\subsection{Relighting LoRA}\label{sec:relight}
Preserving the character's appearance is a crucial feature in character image animation. However, when performing character replacement, a challenge arises because the character and the environment originate from different sources. Strictly maintaining appearance consistency can lead to a mismatch between the animated character's lighting and color tone and those of the new environment, which compromises the realism of the final result.
Therefore, for Replacement Mode, we introduce an auxiliary Relighting LoRA \cite{lora}. This module allows for further adjustments to the character's lighting and color tone during replacement, enabling it to adapt to the new environment. The Relighting LoRA is applied exclusively to the self-attention and cross-attention layers within the DiT blocks.
To train this LoRA, we construct specific data pairs. For a reference image sampled from a video clip, we first segment and crop the character from the original image. We then use IC-Light \cite{iclight} to synthesize the character onto a new, random background. As illustrated in Figure \ref{fig:iclight}, leveraging IC-Light's capabilities, the character's lighting and color tone are influenced by the new background, creating a discrepancy with the original video sequence. This newly synthesized image is then used as the reference, allowing the Relighting LoRA to learn the ability to perform lighting and color adjustments.
When augmented with the Relighting LoRA, \textit{Wan-Animate} can produce a better environmental fusion for the replaced character while simultaneously preserving its identity.

\subsection{Training}
The training process of \textit{Wan-Animate} is divided into the following stages:

\textbf{Body Control Training. }
We first focus exclusively on training the model for Animation Mode. In this stage, conditioning is limited to the body control signal, with no facial signal injection. The goal is for the model to quickly learn our modified input paradigm (i.e., the specific configurations for the reference image and temporal images) and to master the alignment with the body control signal.

\textbf{Face Control Training. }
Next, we introduce facial signal injection. Building upon the model from Stage 1, we integrate the Face Adapter and Face Block modules. To accelerate training, we initialize a portion of their parameters using the pre-trained encoder weights from \cite{lia}. This stage primarily utilizes portrait data, as facial motion is the dominant dynamic in such videos, allowing for a focused learning of expression-driven animation. We also use facial landmarks to identify head, eye, and mouth regions, applying a higher loss weight to these areas to enhance their fidelity.

\textbf{Joint Control Training. }
Here, we combine the Face Adapter and Face Block modules from Stage 2 with the main model trained in Stage 1, and perform joint control training on the full dataset. Our experiments show that the standalone face module already possesses strong expression-driving capabilities, enabling the full model to converge rapidly.

\textbf{Joint Mode Training. }
In this stage, we adapt the training data to include formats for both Animation Mode and Replacement Mode. Given the model's established animation capabilities and the compatibility of our input formulation with Wan-I2V's pre-training, this transition is remarkably smooth.

\textbf{Relighting LoRA Training. }
Finally, we exclusively train the relighting capability for the Replacement Mode by applying the Relighting LoRA. The detailed methodology for this stage is described in Section \ref{sec:relight}.

\subsection{Inference}

\textbf{Pose Retargeting. }
During inference, the characters in the provided image and reference video often have different identities. Due to disparities in bone proportions and relative size, for Animation Mode, we perform pose retargeting on the skeletons extracted from the reference video. This involves calculating the length ratio of each corresponding limb between the two characters and adjusting the target pose's bone lengths to match the character in the source image. Additionally, the pose is translated to align with the character's position in the image. The reference point for this translation is determined by the framing of the shot (e.g., feet for full-body and neck for half-body or portraits). We will open-source a simplified version our retargeting pipeline.
Since we use a 2D skeleton, the character's posture can affect the accuracy of the calculated bone lengths. To mitigate this, we provide an auxiliary solution. Specifically, we use the image editing model \cite{qwenimage,fluxkontext} to edit the characters in both the reference and driving images into a standard T-pose. The scaling factors are then calculated based on the bone lengths from these edited T-pose images. In most scenarios, this approach leads to more accurate retargeting.
In Replacement Mode, given that the character may have specific interactions with the environment, we aim to avoid disrupting these relationships. Therefore, we do not recommend using pose retargeting during character replacement. This, however, introduces a limitation for certain use cases, such as replacing characters with significant body shape differences, which may result in some deformation.

\textbf{Long Video. }
For long video generation, we adopt an iterative generation approach. Specifically, for the first segment, we concatenate only the reference latent and the noise latents. After generating the video result for this segment, we select its last few frames to serve as the temporal guidance for the subsequent segment. The generation of all subsequent segments then involves a concatenation of the reference latent, the temporal latents, and the new noise latents.
Based on practical usage, we typically use one or two latents as temporal guidance, corresponding to 1 or 5 frames, respectively. After the denoising process is complete for each segment, we discard the portions corresponding to the reference latent and the temporal guidance latents. The remaining generated content is then concatenated to form the final long video.

%% file: content/implementation.tex
\section{Implementation}

\subsection{Data Construction}
We collected a large dataset of human-centric videos, covering activities such as speaking, facial expressions, and body movements. We implemented quality measures \cite{dover,unimatch,improved-aesthetic-predictor} similar to those required for general video generation. To ensure identity consistency during training, we verified that each video clip features only a single, consistent character.
We extracted skeleton information for each character, which serves a dual purpose: first, as the motion signal annotation, and second, as a criterion for filtering videos based on character behavior. For the character replacement task, we use the annotated skeletons to track the character and then extract the corresponding character masks using SAM2 \cite{sam2}.
Additionally, we used the QwenVL2.5-72B \cite{Qwen2.5-VL} model to generate textual descriptions for each video to support the post-training requirements of Wan. While \textit{Wan-Animate} supports a degree of textual control, the motion signal is the dominant control factor, making text control a non-core feature. In practice, we recommend using a default text prompt.

\subsection{Parallel Strategy}
Our training process involves loading multiple models: DiT, T5 \cite{t5}, VAE, and CLIP \cite{clip}. For the memory-intensive models, DiT and T5, we employ Fully Sharded Data Parallelism (FSDP) \cite{fsdp} to reduce the per-GPU memory footprint. The remaining models are trained using standard Data Parallelism (DP).
For the DiT model specifically, we also utilize a Context Parallelism scheme, which combines RingAttention and Ulysses \cite{usp} to enable parallel training. This approach further reduces memory consumption and accelerates training speed.
For the frame-wise facial feature extraction within the Face Adapter, we parallelize the computation within each Ulysses group by treating the facial frames from a single video clip as a batch and processing them concurrently.

\subsection{Details}

\textit{Wan-Animate} supports arbitrary output resolutions. In Animation Mode, the output aspect ratio conforms to that of the input character image. In Replacement Mode, it conforms to the reference video's aspect ratio. The final inference resolution is determined based on the total number of video tokens after patchify. For example, we first calculate a target token count based on a standard resolution like $1280\times720$. Then, for a given aspect ratio, we select the resolution that yields a token count closest to this target.
Each inference segment consists of 78 frames. One frame is statically reserved for the character image. Of the remaining 77 frames, for any segment other than the first, 1 or 5 frames are used as temporal reference frames, sourced from the end of the preceding segment.
To maintain high inference efficiency, classifier-free guidance (CFG) is disabled by default. However, in scenarios where finer control over facial expression is desired, CFG can be optionally enabled for the face conditioning input to adjust the reenactment effect.

%% file: content/experiments.tex
\section{Experiments}

\begin{table*}[tb]
  \centering
  \label{tab:quantitative_comp}
  \setlength{\tabcolsep}{5pt}
  \begin{tabular}{lccc@{\hspace{1.25cm}}lccc}
    \toprule
    Method & SSIM$\uparrow$ & LPIPS$\downarrow$ & FVD$\downarrow$ & Method &  SSIM$\uparrow$ & LPIPS$\downarrow$ & FVD$\downarrow$\\
    \midrule
    Moore-AA & 0.761 & 0.288 & 170.07 & LivePortrait & 0.811 & 0.231 & 118.67 \\
    Champ & 0.749 & 0.297 & 177.64 &  AniPortrait & 0.791 & 0.252 & 135.08\\
    MicmicMotion & 0.742 & 0.307 & 184.71 & Emoji & 0.803 & 0.244 & 127.95\\
    Unianimate & 0.787 & 0.271 & 155.03 & X-portrait2 & 0.825 & 0.212 & 98.03\\
    StableAnimator & 0.794 & 0.265 & 147.92 & SkyReel-A1 & 0.821 & 0.231 & 101.45\\
    \textit{Wan-Animate} & \textbf{0.813} & \textbf{0.227} & \textbf{118.65} & \textit{Wan-Animate} & \textbf{0.834} & \textbf{0.205} & \textbf{94.65}\\
    \bottomrule
  \end{tabular}
  \caption{Quantitative comparisons.}
\end{table*}

\subsection{Quantitative Evaluation}

\begin{figure*}[!t]
\centering
\includegraphics[width=\textwidth]{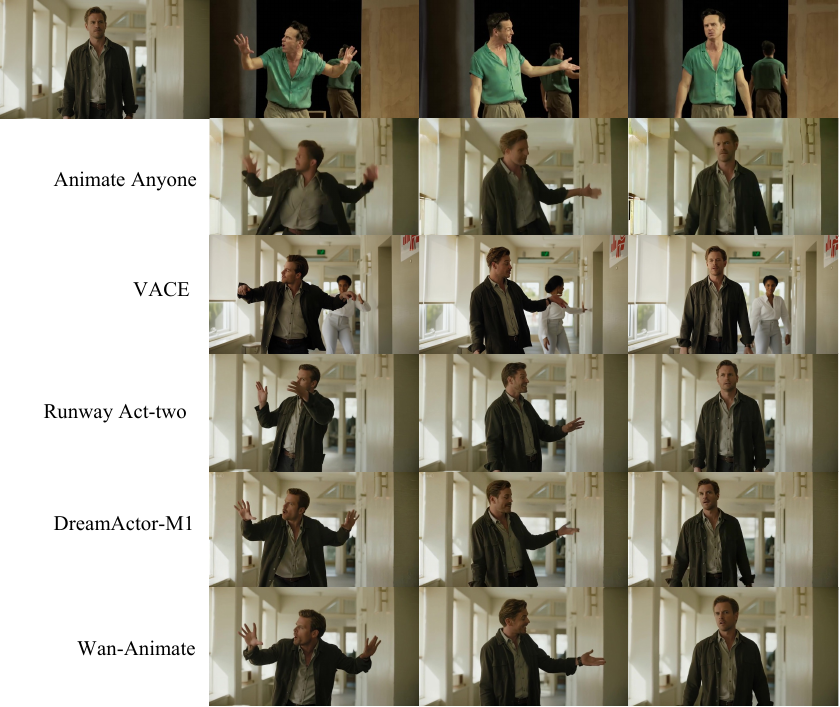}
\caption{Qualitative comparison for Animation Mode.}
\label{fig:com1}
\end{figure*}

We conducted a quantitative comparison with several mainstream open-source character animation frameworks. To facilitate a more comprehensive evaluation, we established our own benchmark for quantitative assessment. The test dataset contains videos of human subjects in various scenarios, featuring different character scales and actions. For the evaluation, we adopted a self-reconstruction task: the first frame of a video is used as the reference image, and the model then reconstructs the video using motion signals from the subsequent frames. We employed several widely-used quantitative metrics, including SSIM \cite{ssim}, LPIPS \cite{lpips}, and FVD \cite{fvd}.
Additionally, we partitioned a subset containing only portraits from our test data to conduct a separate quantitative comparison against specialized facial animation methods. The comparison results are presented in Table \ref{fig:com1}.
Most of the existing open-source frameworks are built upon earlier UNet-based foundation models, which results in certain shortcomings in generation quality, particularly regarding human realism, local details, and temporal smoothness. While recent open-source works based on DiT have improved the performance baseline, they are often limited in their comprehensiveness (e.g., body-driven models lack effective expression reenactment, expression-driven models do not include the body, and support for diverse character types and scales is lacking).
In comparison, \textit{Wan-Animate} performs better than these current open-source works, standing as the most comprehensive and highest-performing open-source model to date.

\begin{figure*}[!t]
\centering
\includegraphics[width=\textwidth]{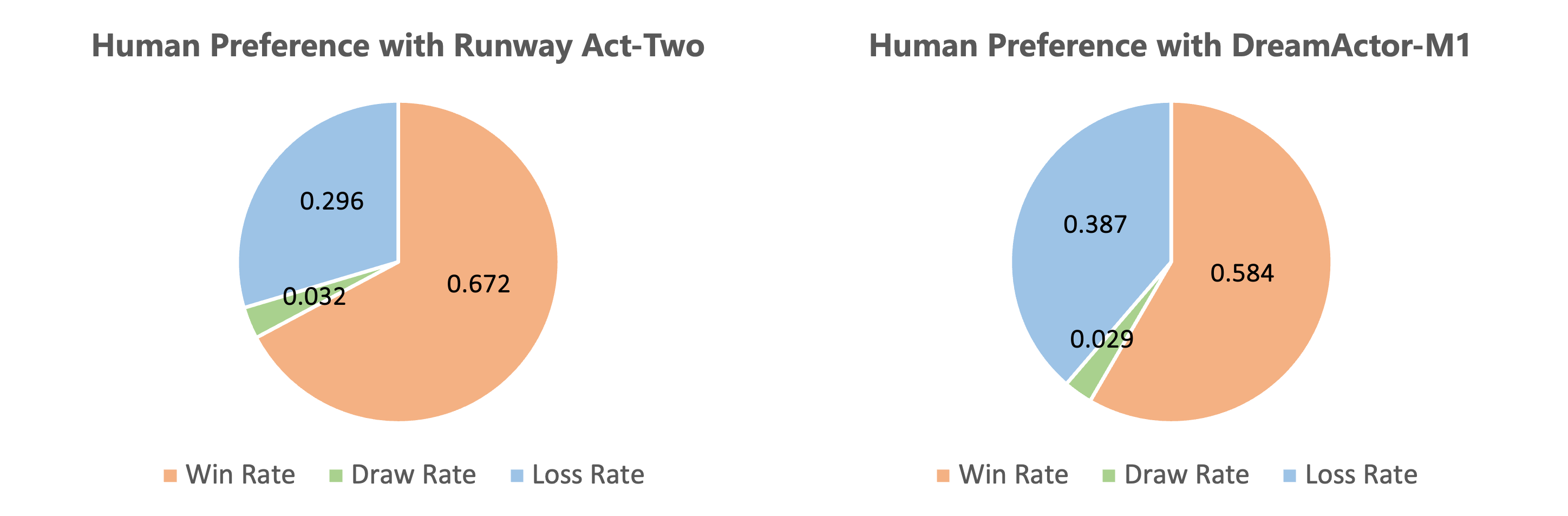}
\caption{
Human evaluation with current SOTA.}
\label{fig:human}
\end{figure*}

\subsection{Human Evaluation}

Currently, the solutions that most closely resemble \textit{Wan-Animate} in terms of both functionality and performance are primarily closed-source: Runway's Act-two \cite{acttwo} and Bytedance's DreamActor-M1 \cite{dreamactorm1}. Compared to existing open-source alternatives, these two proprietary solutions represent the state-of-the-art in character animation in the industry.
We compare \textit{Wan-Animate} with these two methods to demonstrate its superiority. Since conventional quantitative reconstruction metrics may not accurately reflect perceptual differences when the results are of high quality, we employ a cross-ID animation setup and conduct a user study for this comparison.
Each data pair in our evaluation set consists of a driving video and a different character image. After generating the results, we invited 20 participants for a subjective evaluation. Specifically, we presented two generated videos side-by-side in an anonymous fashion (one from \textit{Wan-Animate}, one from a competing method) and asked participants to choose their preferred result. Their preference was based on a comprehensive consideration of video generation quality, overall identity consistency, motion accuracy, and expression accuracy.
The results of the user study are shown in Figure \ref{fig:human}, which clearly indicates that \textit{Wan-Animate} achieved a superior outcome. We believe that the open-sourcing of \textit{Wan-Animate} will raise the performance baseline for open-source models in this domain, contributing to the application and long-term development of this technology.

\begin{figure*}[!t]
\centering
\includegraphics[width=\textwidth]{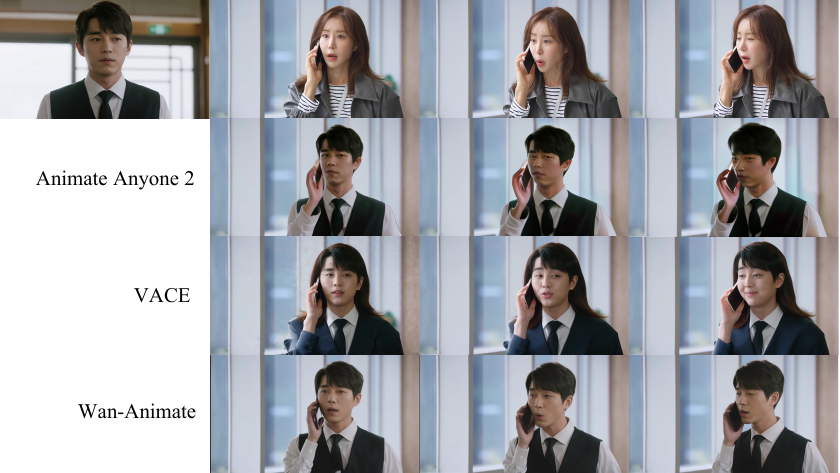}
\caption{Qualitative comparison for Replacement Mode.}
\label{fig:com2}
\end{figure*}

\subsection{Qualitative Evaluation}

In this section, we present a visual comparison of our results.

\textbf{Animation Mode. }
We compare \textit{Wan-Animate} with Animate Anyone, VACE, Runway Act-two, and Dreamactor-M1. As can be seen in Figure \ref{fig:com1}: due to the limitations of its base model, Animate Anyone exhibits significantly lower generation quality. VACE, being a general-purpose controllable video generation model, shows instability in character animation tasks. Runway Act-two struggles significantly with capturing relatively complex motions. DreamActor-M1 tends to have slightly lower quality in local details and overall image fidelity. In comparison, \textit{Wan-Animate} demonstrates a more comprehensive and stable performance overall.

\textbf{Replacement Mode. }
We compare \textit{Wan-Animate} with Animate Anyone 2 and VACE. As shown in Figure \ref{fig:com2}: Animate Anyone 2 also suffers from insufficient generation quality, again likely due to its base model. VACE has issues with identity consistency. Furthermore, its general-purpose nature makes it highly dependent on parameter tuning, resulting in a higher barrier to entry. In contrast, \textit{Wan-Animate} is much more user-friendly and performs better in character replacement.

\begin{figure*}[!t]
\centering
\includegraphics[width=0.75\textwidth]{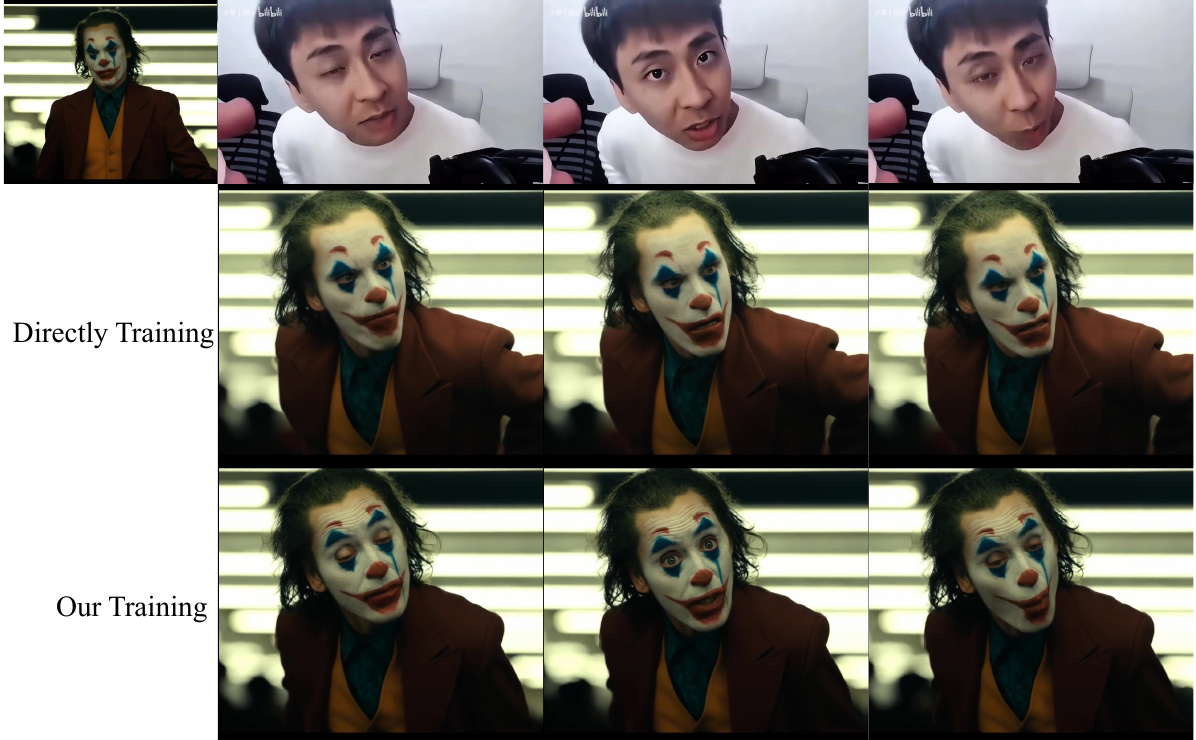}
\caption{Ablation study on Face Adapter Training.}
\label{fig:aba1}
\end{figure*}

\begin{figure*}[!t]
\centering
\includegraphics[width=\textwidth]{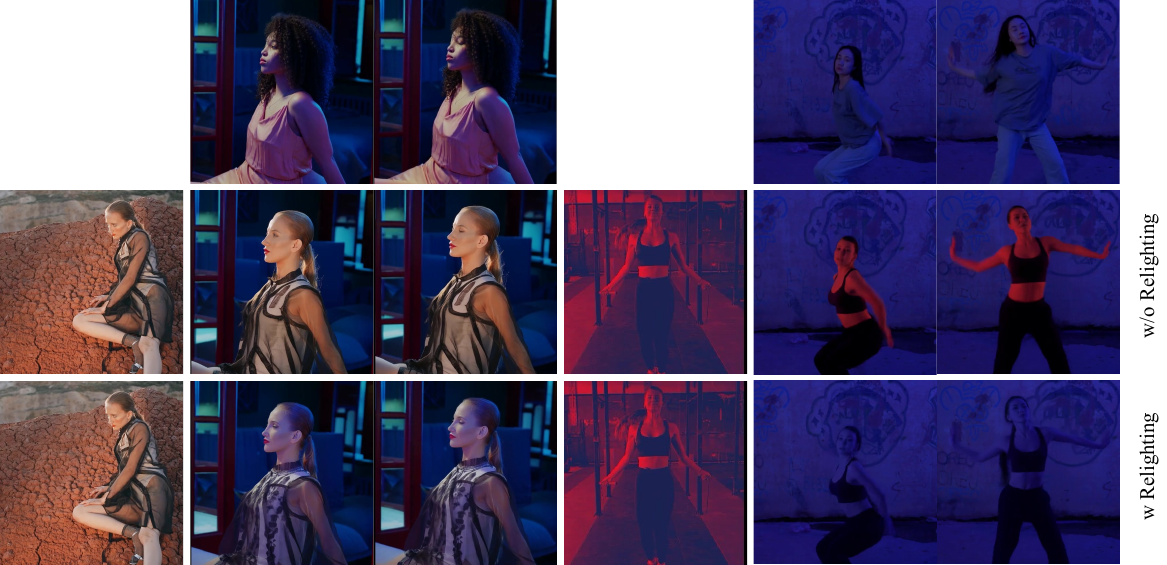}
\caption{Ablation study of Relighting LoRA.}
\label{fig:aba2}
\end{figure*}

\subsection{Ablation Study}
\textbf{Ablation Study on Face Adapter Training. }
Our training scheme employs a progressive pipeline: we first train for body control, then for facial expressions, and finally train them jointly. This process involves specific data usage and training techniques at each stage. This scheme is highly beneficial for the convergence of the face adapter.
To validate its effectiveness, we conduct an ablation study. The baseline for comparison involves training the entire control module jointly on all data from the start. The results are shown in Figure \ref{fig:aba1}. We observe that in the baseline experiment, the expression driving is inaccurate, and the model struggles to converge properly.
We believe this is because body motion is more complex; learning to align the body first facilitates the subsequent learning of expressions. Furthermore, since the face generally occupies a small portion of the frame in typical data, training the expression module on portrait data, where the face is prominent, significantly accelerates its convergence.

\textbf{Effect of Relighting LoRA. }
In Replacement Mode, we train the Relighting LoRA on specifically constructed data to achieve better integration of the character with the environment in terms of lighting and color tone. We conducted an ablation study to verify its effect.
Figure \ref{fig:aba2} shows a comparison of the results with and without the Relighting LoRA. As can be seen, without the LoRA, the character's lighting and color tone in the generated video maintain a strong consistency with the reference image. However, this can appear incongruous when integrated into the new environment. Therefore, the Relighting LoRA adds a degree of flexible adaptability on top of the strong consistency requirement of the character animation task.
With the Relighting LoRA, the fusion of the character and the environment becomes more realistic and harmonious. Critically, this is achieved without breaking the character's perceptual identity.

\begin{figure*}[!t]
\centering
\includegraphics[width=\textwidth]{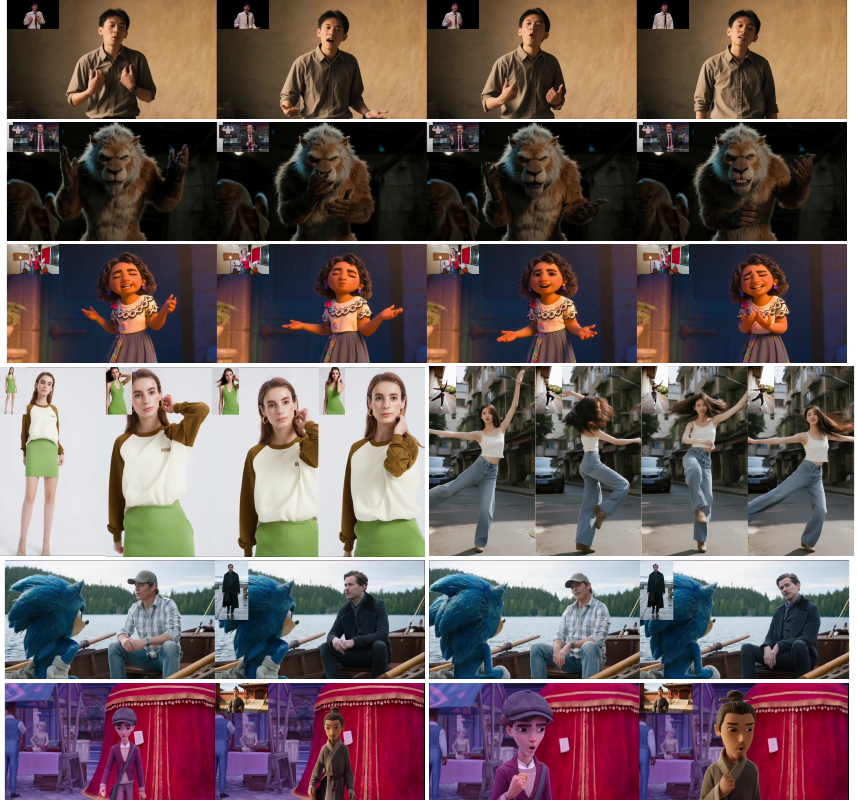}
\caption{Qualitative Results for various applications.}
\label{fig:quali}
\end{figure*}

\subsection{More Qualitative Results}

In Figure \ref{fig:quali}, we showcase a variety of results generated by \textit{Wan-Animate}, demonstrating its wide range of potential applications.
\textbf{Performance Reenactment}: \textit{Wan-Animate} allows a specified person to precisely replicate the performance of a character in a source video, enabling the recreation of classic performance scenes.
\textbf{Cross-Style Transfer}: The model can robustly transfer a real person's performance to various types of characters, which is highly beneficial for filmmaking and animation.
\textbf{Complex Motion Synthesis}: \textit{Wan-Animate} can replicate dance routines and other special actions, facilitating content creation for short-form entertainment videos.
\textbf{Dynamic Camera Movement}: The model can generate character actions that include camera movements, showing its value in advertisement production.
\textbf{Character Replacement}: Furthermore, \textit{Wan-Animate}'s robust character replacement capability facilitates applications such as re-imagining scenes from films and TV series or editing characters in commercial photography and advertising.

%% file: main.bbl
\begin{thebibliography}{58}
\providecommand{\natexlab}[1]{#1}
\providecommand{\url}[1]{\texttt{#1}}
\expandafter\ifx\csname urlstyle\endcsname\relax
  \providecommand{\doi}[1]{doi: #1}\else
  \providecommand{\doi}{doi: \begingroup \urlstyle{rm}\Url}\fi

\bibitem[Bai et~al.(2025)Bai, Chen, Liu, Wang, Ge, Song, Dang, Wang, Wang, Tang, Zhong, Zhu, Yang, Li, Wan, Wang, Ding, Fu, Xu, Ye, Zhang, Xie, Cheng, Zhang, Yang, Xu, and Lin]{Qwen2.5-VL}
Shuai Bai, Keqin Chen, Xuejing Liu, Jialin Wang, Wenbin Ge, Sibo Song, Kai Dang, Peng Wang, Shijie Wang, Jun Tang, Humen Zhong, Yuanzhi Zhu, Mingkun Yang, Zhaohai Li, Jianqiang Wan, Pengfei Wang, Wei Ding, Zheren Fu, Yiheng Xu, Jiabo Ye, Xi~Zhang, Tianbao Xie, Zesen Cheng, Hang Zhang, Zhibo Yang, Haiyang Xu, and Junyang Lin.
\newblock Qwen2.5-vl technical report.
\newblock \emph{arXiv preprint arXiv:2502.13923}, 2025.

\bibitem[Blattmann et~al.(2023)Blattmann, Dockhorn, Kulal, Mendelevitch, Kilian, Lorenz, Levi, English, Voleti, Letts, et~al.]{svd}
Andreas Blattmann, Tim Dockhorn, Sumith Kulal, Daniel Mendelevitch, Maciej Kilian, Dominik Lorenz, Yam Levi, Zion English, Vikram Voleti, Adam Letts, et~al.
\newblock Stable video diffusion: Scaling latent video diffusion models to large datasets.
\newblock \emph{arXiv preprint arXiv:2311.15127}, 2023.

\bibitem[Brooks et~al.(2024)Brooks, Peebles, Holmes, DePue, Guo, Jing, Schnurr, Taylor, Luhman, Luhman, et~al.]{sora}
Tim Brooks, Bill Peebles, Connor Holmes, Will DePue, Yufei Guo, Li~Jing, David Schnurr, Joe Taylor, Troy Luhman, Eric Luhman, et~al.
\newblock Video generation models as world simulators.
\newblock \emph{OpenAI Blog}, 1\penalty0 (8):\penalty0 1, 2024.

\bibitem[Chang et~al.(2023)Chang, Shi, Gao, Xu, Fu, Song, Yan, Zhu, Yang, and Soleymani]{magicpose}
Di~Chang, Yichun Shi, Quankai Gao, Hongyi Xu, Jessica Fu, Guoxian Song, Qing Yan, Yizhe Zhu, Xiao Yang, and Mohammad Soleymani.
\newblock Magicpose: Realistic human poses and facial expressions retargeting with identity-aware diffusion.
\newblock In \emph{Forty-first International Conference on Machine Learning}, 2023.

\bibitem[Drobyshev et~al.(2024)Drobyshev, Casademunt, Vougioukas, Landgraf, Petridis, and Pantic]{emoportraits}
Nikita Drobyshev, Antoni~Bigata Casademunt, Konstantinos Vougioukas, Zoe Landgraf, Stavros Petridis, and Maja Pantic.
\newblock Emoportraits: Emotion-enhanced multimodal one-shot head avatars.
\newblock In \emph{Proceedings of the IEEE/CVF Conference on Computer Vision and Pattern Recognition}, pp.\  8498--8507, 2024.

\bibitem[Fang \& Zhao(2024)Fang and Zhao]{usp}
Jiarui Fang and Shangchun Zhao.
\newblock Usp: A unified sequence parallelism approach for long context generative ai, 2024.
\newblock URL \url{https://arxiv.org/abs/2405.07719}.

\bibitem[Gan et~al.(2025)Gan, Ren, Zhang, Ye, Xie, Yin, Yuan, Peng, and Zhu]{humandit}
Qijun Gan, Yi~Ren, Chen Zhang, Zhenhui Ye, Pan Xie, Xiang Yin, Zehuan Yuan, Bingyue Peng, and Jianke Zhu.
\newblock Humandit: Pose-guided diffusion transformer for long-form human motion video generation.
\newblock \emph{arXiv preprint arXiv:2502.04847}, 2025.

\bibitem[Guo et~al.(2024)Guo, Zhang, Liu, Zhong, Zhang, Wan, and Zhang]{liveportrait}
Jianzhu Guo, Dingyun Zhang, Xiaoqiang Liu, Zhizhou Zhong, Yuan Zhang, Pengfei Wan, and Di~Zhang.
\newblock Liveportrait: Efficient portrait animation with stitching and retargeting control.
\newblock \emph{arXiv preprint arXiv:2407.03168}, 2024.

\bibitem[Guo et~al.(2023)Guo, Yang, Rao, Wang, Qiao, Lin, and Dai]{animatediff}
Yuwei Guo, Ceyuan Yang, Anyi Rao, Yaohui Wang, Yu~Qiao, Dahua Lin, and Bo~Dai.
\newblock Animatediff: Animate your personalized text-to-image diffusion models without specific tuning.
\newblock \emph{arXiv preprint arXiv:2307.04725}, 2023.

\bibitem[Ho et~al.(2020)Ho, Jain, and Abbeel]{ddpm}
Jonathan Ho, Ajay Jain, and Pieter Abbeel.
\newblock Denoising diffusion probabilistic models.
\newblock \emph{Advances in neural information processing systems}, 33:\penalty0 6840--6851, 2020.

\bibitem[Hu et~al.(2022)Hu, Shen, Wallis, Allen-Zhu, Li, Wang, Wang, Chen, et~al.]{lora}
Edward~J Hu, Yelong Shen, Phillip Wallis, Zeyuan Allen-Zhu, Yuanzhi Li, Shean Wang, Lu~Wang, Weizhu Chen, et~al.
\newblock Lora: Low-rank adaptation of large language models.
\newblock \emph{ICLR}, 1\penalty0 (2):\penalty0 3, 2022.

\bibitem[Hu(2024)]{aa}
Li~Hu.
\newblock Animate anyone: Consistent and controllable image-to-video synthesis for character animation.
\newblock In \emph{Proceedings of the IEEE/CVF Conference on Computer Vision and Pattern Recognition}, pp.\  8153--8163, 2024.

\bibitem[Hu et~al.(2025)Hu, Wang, Shen, Gao, Meng, Zhuo, Zhang, Zhang, and Bo]{aa2}
Li~Hu, Guangyuan Wang, Zhen Shen, Xin Gao, Dechao Meng, Lian Zhuo, Peng Zhang, Bang Zhang, and Liefeng Bo.
\newblock Animate anyone 2: High-fidelity character image animation with environment affordance.
\newblock \emph{arXiv preprint arXiv:2502.06145}, 2025.

\bibitem[Jiang et~al.(2025)Jiang, Han, Mao, Zhang, Pan, and Liu]{vace}
Zeyinzi Jiang, Zhen Han, Chaojie Mao, Jingfeng Zhang, Yulin Pan, and Yu~Liu.
\newblock Vace: All-in-one video creation and editing.
\newblock \emph{arXiv preprint arXiv:2503.07598}, 2025.

\bibitem[Ki et~al.(2024)Ki, Min, and Chae]{float}
Taekyung Ki, Dongchan Min, and Gyeongsu Chae.
\newblock Float: Generative motion latent flow matching for audio-driven talking portrait.
\newblock \emph{arXiv preprint arXiv:2412.01064}, 2024.

\bibitem[Kingma \& Welling(2014)Kingma and Welling]{vae}
Diederik~P. Kingma and Max Welling.
\newblock Auto-encoding variational bayes.
\newblock In Yoshua Bengio and Yann LeCun (eds.), \emph{2nd International Conference on Learning Representations, {ICLR} 2014, Banff, AB, Canada, April 14-16, 2014, Conference Track Proceedings}, 2014.
\newblock URL \url{http://arxiv.org/abs/1312.6114}.

\bibitem[Kong et~al.(2025)Kong, Tian, Zhang, Min, Dai, Zhou, Xiong, Li, Wu, Zhang, Wu, Lin, Yuan, Long, Wang, Wang, Li, Huang, Yang, Tan, Wang, Song, Bai, Wu, Xue, Wang, Wang, Liu, Li, Li, Wang, Yu, Deng, Li, Chen, Cui, Peng, Yu, He, Xu, Zhou, Xu, Tao, Lu, Liu, Zhou, Wang, Yang, Wang, Liu, Jiang, and Zhong]{hunyuanvideo}
Weijie Kong, Qi~Tian, Zijian Zhang, Rox Min, Zuozhuo Dai, Jin Zhou, Jiangfeng Xiong, Xin Li, Bo~Wu, Jianwei Zhang, Kathrina Wu, Qin Lin, Junkun Yuan, Yanxin Long, Aladdin Wang, Andong Wang, Changlin Li, Duojun Huang, Fang Yang, Hao Tan, Hongmei Wang, Jacob Song, Jiawang Bai, Jianbing Wu, Jinbao Xue, Joey Wang, Kai Wang, Mengyang Liu, Pengyu Li, Shuai Li, Weiyan Wang, Wenqing Yu, Xinchi Deng, Yang Li, Yi~Chen, Yutao Cui, Yuanbo Peng, Zhentao Yu, Zhiyu He, Zhiyong Xu, Zixiang Zhou, Zunnan Xu, Yangyu Tao, Qinglin Lu, Songtao Liu, Daquan Zhou, Hongfa Wang, Yong Yang, Di~Wang, Yuhong Liu, Jie Jiang, and Caesar Zhong.
\newblock Hunyuanvideo: A systematic framework for large video generative models, 2025.
\newblock URL \url{https://arxiv.org/abs/2412.03603}.

\bibitem[Labs et~al.(2025)Labs, Batifol, Blattmann, Boesel, Consul, Diagne, Dockhorn, English, English, Esser, et~al.]{fluxkontext}
Black~Forest Labs, Stephen Batifol, Andreas Blattmann, Frederic Boesel, Saksham Consul, Cyril Diagne, Tim Dockhorn, Jack English, Zion English, Patrick Esser, et~al.
\newblock Flux. 1 kontext: Flow matching for in-context image generation and editing in latent space.
\newblock \emph{arXiv preprint arXiv:2506.15742}, 2025.

\bibitem[Loper et~al.(2023)Loper, Mahmood, Romero, Pons-Moll, and Black]{smpl}
Matthew Loper, Naureen Mahmood, Javier Romero, Gerard Pons-Moll, and Michael~J Black.
\newblock Smpl: A skinned multi-person linear model.
\newblock In \emph{Seminal Graphics Papers: Pushing the Boundaries, Volume 2}, pp.\  851--866. 2023.

\bibitem[Luo et~al.(2025)Luo, Rong, Wang, Zhang, Hu, and Zhu]{dreamactorm1}
Yuxuan Luo, Zhengkun Rong, Lizhen Wang, Longhao Zhang, Tianshu Hu, and Yongming Zhu.
\newblock Dreamactor-m1: Holistic, expressive and robust human image animation with hybrid guidance.
\newblock \emph{arXiv preprint arXiv:2504.01724}, 2025.

\bibitem[Ma et~al.(2024)Ma, Liu, Wang, Pan, He, Yuan, Zeng, Cai, Shum, Liu, et~al.]{emoji}
Yue Ma, Hongyu Liu, Hongfa Wang, Heng Pan, Yingqing He, Junkun Yuan, Ailing Zeng, Chengfei Cai, Heung-Yeung Shum, Wei Liu, et~al.
\newblock Follow-your-emoji: Fine-controllable and expressive freestyle portrait animation.
\newblock In \emph{SIGGRAPH Asia 2024 Conference Papers}, pp.\  1--12, 2024.

\bibitem[Men et~al.(2024)Men, Yao, Cui, and Bo]{mimo}
Yifang Men, Yuan Yao, Miaomiao Cui, and Liefeng Bo.
\newblock Mimo: Controllable character video synthesis with spatial decomposed modeling.
\newblock \emph{arXiv preprint arXiv:2409.16160}, 2024.

\bibitem[Peebles \& Xie(2023)Peebles and Xie]{dit}
William Peebles and Saining Xie.
\newblock Scalable diffusion models with transformers.
\newblock In \emph{Proceedings of the IEEE/CVF international conference on computer vision}, pp.\  4195--4205, 2023.

\bibitem[Radford et~al.(2021)Radford, Kim, Hallacy, Ramesh, Goh, Agarwal, Sastry, Askell, Mishkin, Clark, et~al.]{clip}
Alec Radford, Jong~Wook Kim, Chris Hallacy, Aditya Ramesh, Gabriel Goh, Sandhini Agarwal, Girish Sastry, Amanda Askell, Pamela Mishkin, Jack Clark, et~al.
\newblock Learning transferable visual models from natural language supervision.
\newblock In \emph{International conference on machine learning}, pp.\  8748--8763. PMLR, 2021.

\bibitem[Raffel et~al.(2020)Raffel, Shazeer, Roberts, Lee, Narang, Matena, Zhou, Li, and Liu]{t5}
Colin Raffel, Noam Shazeer, Adam Roberts, Katherine Lee, Sharan Narang, Michael Matena, Yanqi Zhou, Wei Li, and Peter~J Liu.
\newblock Exploring the limits of transfer learning with a unified text-to-text transformer.
\newblock \emph{Journal of machine learning research}, 21\penalty0 (140):\penalty0 1--67, 2020.

\bibitem[Ravi et~al.(2024)Ravi, Gabeur, Hu, Hu, Ryali, Ma, Khedr, R{\"a}dle, Rolland, Gustafson, et~al.]{sam2}
Nikhila Ravi, Valentin Gabeur, Yuan-Ting Hu, Ronghang Hu, Chaitanya Ryali, Tengyu Ma, Haitham Khedr, Roman R{\"a}dle, Chloe Rolland, Laura Gustafson, et~al.
\newblock Sam 2: Segment anything in images and videos.
\newblock \emph{arXiv preprint arXiv:2408.00714}, 2024.

\bibitem[Rombach et~al.(2022)Rombach, Blattmann, Lorenz, Esser, and Ommer]{ldm}
Robin Rombach, Andreas Blattmann, Dominik Lorenz, Patrick Esser, and Bj{\"o}rn Ommer.
\newblock High-resolution image synthesis with latent diffusion models.
\newblock In \emph{Proceedings of the IEEE/CVF conference on computer vision and pattern recognition}, pp.\  10684--10695, 2022.

\bibitem[Runway(2025)]{acttwo}
Runway.
\newblock Creating with act-two, 2025.
\newblock URL \url{https://help.runwayml.com/hc/en-us/articles/42311337895827-Creating-with-Act-Two}.

\bibitem[Schuhmann(2022)]{improved-aesthetic-predictor}
Christoph Schuhmann.
\newblock improved-aesthetic-predictor.
\newblock \url{https://github.com/christophschuhmann/improved-aesthetic-predictor}, 2022.

\bibitem[Siarohin et~al.(2019)Siarohin, Lathuili{\`e}re, Tulyakov, Ricci, and Sebe]{fomm}
Aliaksandr Siarohin, St{\'e}phane Lathuili{\`e}re, Sergey Tulyakov, Elisa Ricci, and Nicu Sebe.
\newblock First order motion model for image animation.
\newblock \emph{Advances in neural information processing systems}, 32, 2019.

\bibitem[Siarohin et~al.(2021)Siarohin, Woodford, Ren, Chai, and Tulyakov]{mraa}
Aliaksandr Siarohin, Oliver~J Woodford, Jian Ren, Menglei Chai, and Sergey Tulyakov.
\newblock Motion representations for articulated animation.
\newblock In \emph{Proceedings of the IEEE/CVF Conference on Computer Vision and Pattern Recognition}, pp.\  13653--13662, 2021.

\bibitem[Tu et~al.(2025)Tu, Xing, Han, Cheng, Dai, Luo, and Wu]{stableanimator}
Shuyuan Tu, Zhen Xing, Xintong Han, Zhi-Qi Cheng, Qi~Dai, Chong Luo, and Zuxuan Wu.
\newblock Stableanimator: High-quality identity-preserving human image animation.
\newblock In \emph{Proceedings of the Computer Vision and Pattern Recognition Conference}, pp.\  21096--21106, 2025.

\bibitem[Unterthiner et~al.(2018)Unterthiner, Van~Steenkiste, Kurach, Marinier, Michalski, and Gelly]{fvd}
Thomas Unterthiner, Sjoerd Van~Steenkiste, Karol Kurach, Raphael Marinier, Marcin Michalski, and Sylvain Gelly.
\newblock Towards accurate generative models of video: A new metric \& challenges.
\newblock \emph{arXiv preprint arXiv:1812.01717}, 2018.

\bibitem[Wan et~al.(2025)Wan, Wang, Ai, Wen, Mao, Xie, Chen, Yu, Zhao, Yang, Zeng, Wang, Zhang, Zhou, Wang, Chen, Zhu, Zhao, Yan, Huang, Feng, Zhang, Li, Wu, Chu, Feng, Zhang, Sun, Fang, Wang, Gui, Weng, Shen, Lin, Wang, Wang, Zhou, Wang, Shen, Yu, Shi, Huang, Xu, Kou, Lv, Li, Liu, Wang, Zhang, Huang, Li, Wu, Liu, Pan, Zheng, Hong, Shi, Feng, Jiang, Han, Wu, and Liu]{wan2025}
Team Wan, Ang Wang, Baole Ai, Bin Wen, Chaojie Mao, Chen-Wei Xie, Di~Chen, Feiwu Yu, Haiming Zhao, Jianxiao Yang, Jianyuan Zeng, Jiayu Wang, Jingfeng Zhang, Jingren Zhou, Jinkai Wang, Jixuan Chen, Kai Zhu, Kang Zhao, Keyu Yan, Lianghua Huang, Mengyang Feng, Ningyi Zhang, Pandeng Li, Pingyu Wu, Ruihang Chu, Ruili Feng, Shiwei Zhang, Siyang Sun, Tao Fang, Tianxing Wang, Tianyi Gui, Tingyu Weng, Tong Shen, Wei Lin, Wei Wang, Wei Wang, Wenmeng Zhou, Wente Wang, Wenting Shen, Wenyuan Yu, Xianzhong Shi, Xiaoming Huang, Xin Xu, Yan Kou, Yangyu Lv, Yifei Li, Yijing Liu, Yiming Wang, Yingya Zhang, Yitong Huang, Yong Li, You Wu, Yu~Liu, Yulin Pan, Yun Zheng, Yuntao Hong, Yupeng Shi, Yutong Feng, Zeyinzi Jiang, Zhen Han, Zhi-Fan Wu, and Ziyu Liu.
\newblock Wan: Open and advanced large-scale video generative models.
\newblock \emph{arXiv preprint arXiv:2503.20314}, 2025.

\bibitem[Wang et~al.(2025{\natexlab{a}})Wang, Xia, Hu, Wang, Wang, Zheng, and Zhou]{dreamactorh1}
Lizhen Wang, Zhurong Xia, Tianshu Hu, Pengrui Wang, Pengfei Wang, Zerong Zheng, and Ming Zhou.
\newblock Dreamactor-h1: High-fidelity human-product demonstration video generation via motion-designed diffusion transformers.
\newblock \emph{arXiv preprint arXiv:2506.10568}, 2025{\natexlab{a}}.

\bibitem[Wang et~al.(2024)Wang, Zhang, Gao, Wang, Zhou, Zhang, Yan, and Sang]{unianimate}
Xiang Wang, Shiwei Zhang, Changxin Gao, Jiayu Wang, Xiaoqiang Zhou, Yingya Zhang, Luxin Yan, and Nong Sang.
\newblock Unianimate: Taming unified video diffusion models for consistent human image animation.
\newblock \emph{arXiv preprint arXiv:2406.01188}, 2024.

\bibitem[Wang et~al.(2025{\natexlab{b}})Wang, Zhang, Tang, Zhang, Gao, Wang, and Sang]{unianimatedit}
Xiang Wang, Shiwei Zhang, Longxiang Tang, Yingya Zhang, Changxin Gao, Yuehuan Wang, and Nong Sang.
\newblock Unianimate-dit: Human image animation with large-scale video diffusion transformer.
\newblock \emph{arXiv preprint arXiv:2504.11289}, 2025{\natexlab{b}}.

\bibitem[Wang et~al.(2022)Wang, Yang, Bremond, and Dantcheva]{lia}
Yaohui Wang, Di~Yang, Francois Bremond, and Antitza Dantcheva.
\newblock Latent image animator: Learning to animate images via latent space navigation.
\newblock \emph{arXiv preprint arXiv:2203.09043}, 2022.

\bibitem[Wang et~al.(2004)Wang, Bovik, Sheikh, and Simoncelli]{ssim}
Zhou Wang, Alan~C Bovik, Hamid~R Sheikh, and Eero~P Simoncelli.
\newblock Image quality assessment: from error visibility to structural similarity.
\newblock \emph{IEEE transactions on image processing}, 13\penalty0 (4):\penalty0 600--612, 2004.

\bibitem[Wei et~al.(2024)Wei, Yang, and Wang]{aniportrait}
Huawei Wei, Zejun Yang, and Zhisheng Wang.
\newblock Aniportrait: Audio-driven synthesis of photorealistic portrait animation.
\newblock \emph{arXiv preprint arXiv:2403.17694}, 2024.

\bibitem[Wu et~al.(2025)Wu, Li, Zhou, Lin, Gao, Yan, Yin, Bai, Xu, Chen, et~al.]{qwenimage}
Chenfei Wu, Jiahao Li, Jingren Zhou, Junyang Lin, Kaiyuan Gao, Kun Yan, Sheng-ming Yin, Shuai Bai, Xiao Xu, Yilei Chen, et~al.
\newblock Qwen-image technical report.
\newblock \emph{arXiv preprint arXiv:2508.02324}, 2025.

\bibitem[Wu et~al.(2023)Wu, Zhang, Liao, Chen, Hou, Wang, Sun, Yan, and Lin]{dover}
Haoning Wu, Erli Zhang, Liang Liao, Chaofeng Chen, Jingwen~Hou Hou, Annan Wang, Wenxiu~Sun Sun, Qiong Yan, and Weisi Lin.
\newblock Exploring video quality assessment on user generated contents from aesthetic and technical perspectives.
\newblock In \emph{International Conference on Computer Vision (ICCV)}, 2023.

\bibitem[Xie et~al.(2024)Xie, Xu, Song, Wang, Shi, and Luo]{xportrait}
You Xie, Hongyi Xu, Guoxian Song, Chao Wang, Yichun Shi, and Linjie Luo.
\newblock X-portrait: Expressive portrait animation with hierarchical motion attention.
\newblock In \emph{ACM SIGGRAPH 2024 Conference Papers}, pp.\  1--11, 2024.

\bibitem[Xu et~al.(2023)Xu, Zhang, Cai, Rezatofighi, Yu, Tao, and Geiger]{unimatch}
Haofei Xu, Jing Zhang, Jianfei Cai, Hamid Rezatofighi, Fisher Yu, Dacheng Tao, and Andreas Geiger.
\newblock Unifying flow, stereo and depth estimation.
\newblock \emph{IEEE Transactions on Pattern Analysis and Machine Intelligence}, 2023.

\bibitem[Xu et~al.(2024{\natexlab{a}})Xu, Chen, Guo, Yang, Li, Zang, Zhang, Tong, and Guo]{vasa}
Sicheng Xu, Guojun Chen, Yu-Xiao Guo, Jiaolong Yang, Chong Li, Zhenyu Zang, Yizhong Zhang, Xin Tong, and Baining Guo.
\newblock Vasa-1: Lifelike audio-driven talking faces generated in real time.
\newblock \emph{Advances in Neural Information Processing Systems}, 37:\penalty0 660--684, 2024{\natexlab{a}}.

\bibitem[Xu et~al.(2022)Xu, Zhang, Zhang, and Tao]{vitpose}
Yufei Xu, Jing Zhang, Qiming Zhang, and Dacheng Tao.
\newblock Vitpose: Simple vision transformer baselines for human pose estimation, 2022.
\newblock URL \url{https://arxiv.org/abs/2204.12484}.

\bibitem[Xu et~al.(2024{\natexlab{b}})Xu, Zhang, Liew, Yan, Liu, Zhang, Feng, and Shou]{magicanimate}
Zhongcong Xu, Jianfeng Zhang, Jun~Hao Liew, Hanshu Yan, Jia-Wei Liu, Chenxu Zhang, Jiashi Feng, and Mike~Zheng Shou.
\newblock Magicanimate: Temporally consistent human image animation using diffusion model.
\newblock In \emph{Proceedings of the IEEE/CVF Conference on Computer Vision and Pattern Recognition}, pp.\  1481--1490, 2024{\natexlab{b}}.

\bibitem[Xu et~al.(2024{\natexlab{c}})Xu, Huang, Cao, Zhang, Cun, Shuai, Wang, Bao, Li, and Tang]{anchorcrafter}
Ziyi Xu, Ziyao Huang, Juan Cao, Yong Zhang, Xiaodong Cun, Qing Shuai, Yuchen Wang, Linchao Bao, Jintao Li, and Fan Tang.
\newblock Anchorcrafter: Animate cyberanchors saling your products via human-object interacting video generation.
\newblock \emph{arXiv preprint arXiv:2411.17383}, 2024{\natexlab{c}}.

\bibitem[Xu et~al.(2025)Xu, Yu, Zhou, Zhou, Jin, Hong, Ji, Zhu, Cai, Tang, et~al.]{hunyuanportrait}
Zunnan Xu, Zhentao Yu, Zixiang Zhou, Jun Zhou, Xiaoyu Jin, Fa-Ting Hong, Xiaozhong Ji, Junwei Zhu, Chengfei Cai, Shiyu Tang, et~al.
\newblock Hunyuanportrait: Implicit condition control for enhanced portrait animation.
\newblock In \emph{Proceedings of the Computer Vision and Pattern Recognition Conference}, pp.\  15909--15919, 2025.

\bibitem[Yang et~al.(2024)Yang, Teng, Zheng, Ding, Huang, Xu, Yang, Hong, Zhang, Feng, et~al.]{yang2024cogvideox}
Zhuoyi Yang, Jiayan Teng, Wendi Zheng, Ming Ding, Shiyu Huang, Jiazheng Xu, Yuanming Yang, Wenyi Hong, Xiaohan Zhang, Guanyu Feng, et~al.
\newblock Cogvideox: Text-to-video diffusion models with an expert transformer.
\newblock \emph{arXiv preprint arXiv:2408.06072}, 2024.

\bibitem[Zhang et~al.(2025)Zhang, Rao, and Agrawala]{iclight}
Lvmin Zhang, Anyi Rao, and Maneesh Agrawala.
\newblock Scaling in-the-wild training for diffusion-based illumination harmonization and editing by imposing consistent light transport.
\newblock In \emph{The Thirteenth International Conference on Learning Representations}, 2025.

\bibitem[Zhang et~al.(2018)Zhang, Isola, Efros, Shechtman, and Wang]{lpips}
Richard Zhang, Phillip Isola, Alexei~A Efros, Eli Shechtman, and Oliver Wang.
\newblock The unreasonable effectiveness of deep features as a perceptual metric.
\newblock In \emph{Proceedings of the IEEE conference on computer vision and pattern recognition}, pp.\  586--595, 2018.

\bibitem[Zhang et~al.(2024)Zhang, Gu, Wang, Wang, Cheng, Zhu, and Zou]{mimicmotion}
Yuang Zhang, Jiaxi Gu, Li-Wen Wang, Han Wang, Junqi Cheng, Yuefeng Zhu, and Fangyuan Zou.
\newblock Mimicmotion: High-quality human motion video generation with confidence-aware pose guidance.
\newblock \emph{arXiv preprint arXiv:2406.19680}, 2024.

\bibitem[Zhao \& Zhang(2022)Zhao and Zhang]{tpsmm}
Jian Zhao and Hui Zhang.
\newblock Thin-plate spline motion model for image animation.
\newblock In \emph{Proceedings of the IEEE/CVF Conference on Computer Vision and Pattern Recognition}, pp.\  3657--3666, 2022.

\bibitem[Zhao et~al.(2025)Zhao, Xu, Song, Xie, Zhang, Li, Luo, Suo, and Liu]{xportrait2}
Xiaochen Zhao, Hongyi Xu, Guoxian Song, You Xie, Chenxu Zhang, Xiu Li, Linjie Luo, Jinli Suo, and Yebin Liu.
\newblock X-nemo: Expressive neural motion reenactment via disentangled latent attention.
\newblock \emph{arXiv preprint arXiv:2507.23143}, 2025.

\bibitem[Zhao et~al.(2023)Zhao, Gu, Varma, Luo, Huang, Xu, Wright, Shojanazeri, Ott, Shleifer, Desmaison, Balioglu, Damania, Nguyen, Chauhan, Hao, Mathews, and Li]{fsdp}
Yanli Zhao, Andrew Gu, Rohan Varma, Liang Luo, Chien-Chin Huang, Min Xu, Less Wright, Hamid Shojanazeri, Myle Ott, Sam Shleifer, Alban Desmaison, Can Balioglu, Pritam Damania, Bernard Nguyen, Geeta Chauhan, Yuchen Hao, Ajit Mathews, and Shen Li.
\newblock Pytorch fsdp: Experiences on scaling fully sharded data parallel, 2023.
\newblock URL \url{https://arxiv.org/abs/2304.11277}.

\bibitem[Zhou et~al.(2025)Zhou, Wu, Li, Wei, Fan, Chen, Jiang, and Wang]{realisdancedit}
Jingkai Zhou, Yifan Wu, Shikai Li, Min Wei, Chao Fan, Weihua Chen, Wei Jiang, and Fan Wang.
\newblock Realisdance-dit: Simple yet strong baseline towards controllable character animation in the wild.
\newblock \emph{arXiv preprint arXiv:2504.14977}, 2025.

\bibitem[Zhu et~al.(2025)Zhu, Chen, Dai, Dong, Xu, Cao, Yao, Zhu, and Zhu]{champ}
Shenhao Zhu, Junming~Leo Chen, Zuozhuo Dai, Zilong Dong, Yinghui Xu, Xun Cao, Yao Yao, Hao Zhu, and Siyu Zhu.
\newblock Champ: Controllable and consistent human image animation with 3d parametric guidance.
\newblock In \emph{European Conference on Computer Vision}, pp.\  145--162. Springer, 2025.

\end{thebibliography}
